\documentclass[11pt]{article}

\usepackage{acl}
\usepackage{times}
\usepackage{latexsym}
\usepackage[T1]{fontenc}
\usepackage[utf8]{inputenc}
\usepackage{microtype}
\usepackage{inconsolata}
\usepackage{graphicx}
\usepackage{amssymb}
\usepackage{amsmath}
\usepackage{booktabs}
\usepackage{colortbl}
\usepackage{array}
\usepackage{multirow}
\usepackage{algorithm}
\usepackage{algorithmic}
\PassOptionsToPackage{table,dvipsnames}{xcolor}
\usepackage[most]{tcolorbox}
\usepackage{listings}
\usepackage{enumitem}
\usepackage{table-style}
\newcommand{\reproagentScore}{73.7}

\newcommand{\geminiFullScore}{39.7}
\newcommand{\geminiFullMedianScore}{41.8}

\newcommand{\geminiNoReqScore}{25.6}
\newcommand{\geminiNoReqMedianScore}{24.6}

\newcommand{\geminiNoRefScore}{21.6}
\newcommand{\geminiNoRefMedianScore}{21.4}
\newcommand{\geminiNoRefDrop}{18.1}
\newcommand{\geminiNoReqDrop}{14.1}
\newcommand{\deepcodeScore}{73.5}
\newcommand{\basicAgentScore}{19.3}
\newcommand{\iterAgentScore}{20.6}
\newcommand{\aiScientistScore}{30.5}

\newcommand{\system}{\textsc{ReproAgent}}
\definecolor{reproboxbg}{HTML}{F8FAFC}
\definecolor{reproboxframe}{HTML}{2F5D7C}
\definecolor{reproboxsoft}{HTML}{EAF2FF}
\definecolor{repropromptbg}{HTML}{FFF8E8}
\definecolor{repropromptframe}{HTML}{9A6B18}
\definecolor{bestscorebg}{HTML}{FBBDC4}
\definecolor{secondscorebg}{HTML}{FEECEA}

\DeclareRobustCommand{\bestlegend}{\textbf{best}}
\DeclareRobustCommand{\secondlegend}{\underline{second-best}}
\newtcolorbox{detailbox}[2][]{
  enhanced,
  breakable,
  colback=reproboxbg,
  colframe=reproboxframe,
  coltitle=black,
  fonttitle=\bfseries,
  title={#2},
  boxrule=0.6pt,
  arc=1.5mm,
  left=1mm,
  right=1mm,
  top=1mm,
  bottom=1mm,
  #1
}
\newtcblisting{repocode}[2][]{
  enhanced,
  breakable,
  listing only,
  colback=reproboxsoft,
  colframe=reproboxframe,
  coltitle=black,
  fonttitle=\bfseries,
  title={#2},
  boxrule=0.6pt,
  arc=1.5mm,
  left=1mm,
  right=1mm,
  top=1mm,
  bottom=1mm,
  listing options={
    basicstyle=\ttfamily\footnotesize,
    breaklines=true,
    columns=fullflexible,
    keepspaces=true,
    showstringspaces=false
  },
  #1
}
\newtcblisting{promptbox}[2][]{
  enhanced,
  breakable,
  listing only,
  colback=repropromptbg,
  colframe=repropromptframe,
  coltitle=black,
  fonttitle=\bfseries,
  title={#2},
  boxrule=0.6pt,
  arc=1.5mm,
  left=1mm,
  right=1mm,
  top=1mm,
  bottom=1mm,
  listing options={
    basicstyle=\ttfamily\footnotesize,
    breaklines=true,
    columns=fullflexible,
    keepspaces=true,
    showstringspaces=false
  },
  #1
}

\title{ReproAgent: Contract-Guided Paper-to-Code Reproduction}

\author{
  \textbf{Xue Hu}$^{*}$\textsuperscript{1},
  \textbf{Zewei Pan}$^{*}$\textsuperscript{2},
  \textbf{Zhongyuan Wang}\textsuperscript{1},
  \textbf{Zhou Liu}\textsuperscript{3},
  \textbf{Zeli Su}\textsuperscript{4},
  \textbf{Wentao Zhang}$^{\dagger}$\textsuperscript{3,5,6} \\
  \textsuperscript{1}Beihang University, \textsuperscript{2}Shanghai Jiao Tong University, \textsuperscript{3}Peking University \\
  \textsuperscript{4}Minzu University of China, \textsuperscript{5}Zhongguancun Academy \\
  \textsuperscript{6}Beijing Key Laboratory of Data Intelligence and Security (Peking University) \\
  \small{\texttt{kernel@buaa.edu.cn}, \texttt{pzp0057@sjtu.edu.cn},
  \texttt{zywang111@buaa.edu.cn}} \\
  \small{\texttt{zhouliu25@stu.pku.edu.cn}, \texttt{rickamorty@muc.edu.cn},
  \texttt{wentao.zhang@pku.edu.cn}} \\
  \small{$^{*}$Equal contribution.\quad $^{\dagger}$Corresponding author.}
}

\begin{document}
\maketitle

\begin{abstract}
Paper-to-code reproduction asks scientific AI agents to turn research papers into executable repositories that preserve the paper's method, protocol and artifacts. This is difficult because the specification is split: explicit paper content such as algorithms, metrics and artifacts is often lost across long agent trajectories, while implicit details such as framework defaults and conventions inherited from related work are absent from the paper. We introduce \system{}, a four-stage Prepare--Plan--Generate--Repair pipeline built around a persistent implementation contract with two channels: an implementation-requirement channel that turns paper snippets into code obligations, and a reference-evidence channel that retrieves content and structure evidence from related repositories. Both are bound to work packages, projected into file-level contracts, and consumed across generation and repair. On PaperBench Code-Dev~\citep{paperbench2025}, ReproAgent reaches the highest mean score among same-backbone scaffolds under both Claude-Sonnet-4.5 and Gemini-3-Flash. End-to-end channel ablations and per-paper cases support the contribution of both channels. Code and experimental artifacts are publicly available.\footnote{\url{https://github.com/kernel-14/ReproAgent}.}
\end{abstract}

\section{Introduction}
\label{sec:introduction}
 
\begin{figure*}[t!]
  \centering
  \includegraphics[width=0.94\textwidth]{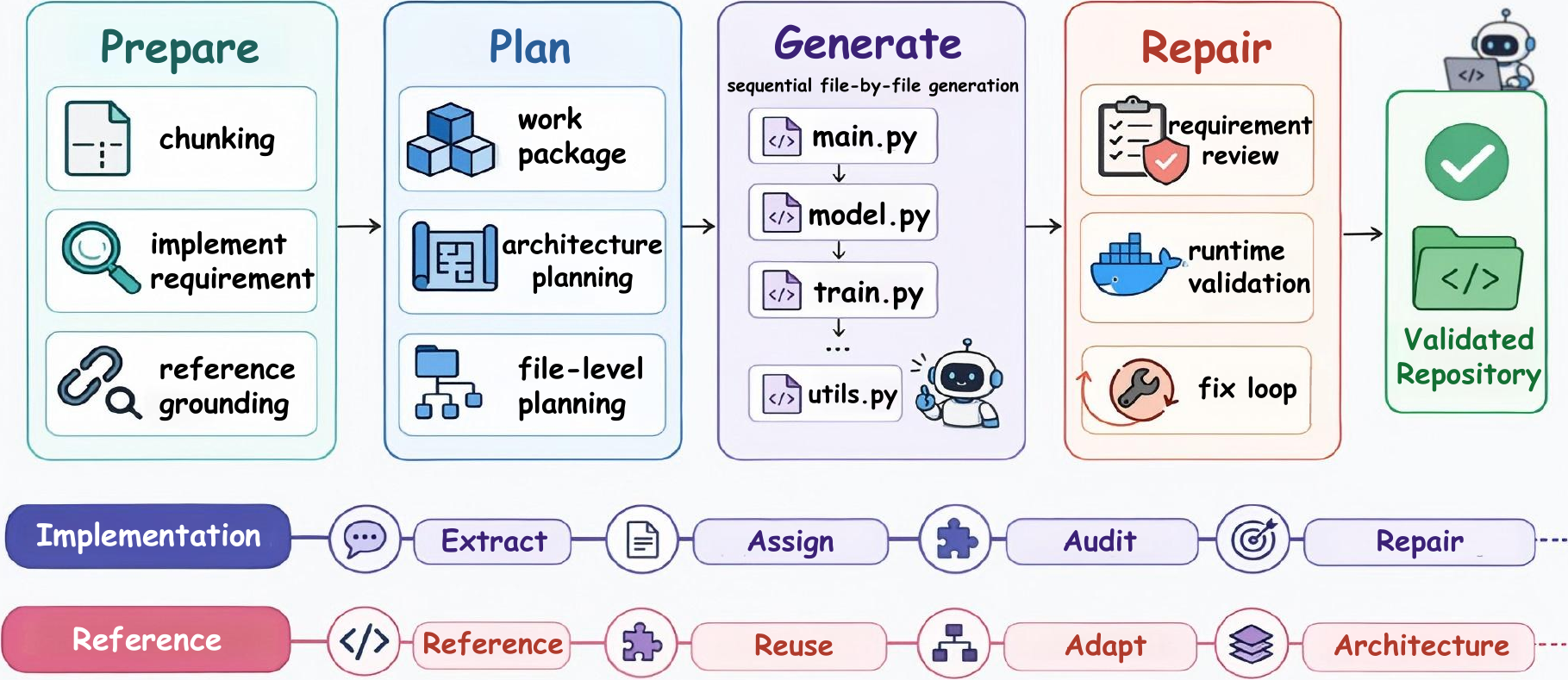}
  \caption{Overview of \system{}. A four-stage pipeline, Prepare, Plan, Generate, and Repair, carries an implementation-requirement channel and a reference-evidence channel into a shared file-level implementation contract. The implementation-requirement channel preserves paper-derived code obligations across generation and repair, while the reference-evidence channel supplies package-local content and structure evidence from related repositories.}
  \label{fig:pipeline}
\end{figure*}

Large language models are increasingly used to assist scientific research, but a long-standing reproducibility crisis persists~\citep{baker2016,pineau2021reproducibility}.
Systematic reviews of NLP reproduction studies report that only a small fraction of attempted reproductions recover the original scores~\citep{belz2021reproducibility}, and even strong LLM judges detect only 46.7\% of real-world discrepancies between papers and their accompanying code~\citep{baumgartner2026scicoqa}.
Together, faster method production and persistent reproducibility gaps make it harder to validate new methods, compare baselines fairly, and build on prior work.
An automated framework that turns each new paper into a faithful, runnable repository, rather than merely executable code, would make paper-specific methods, protocols, and artifacts easier to inspect and reuse.

Recent paper-to-code agents take a step in this direction~\citep{papercoder2026,scireproducer2025,autoreproduce2025,repro2025}, but a substantial fidelity gap remains.
A recurring failure mode is runnable but unfaithful code: a generated repository may import correctly, pass smoke tests, and even produce plausible experimental results, while still dropping or misimplementing key paper-specific details, such as substituting a generic training loop for the paper's algorithm, omitting the decisive metric or artifact, or altering a loss term.
Runtime success alone is therefore an insufficient proxy for faithful paper reproduction.

We identify a split-specification problem behind these failures. A paper states some implementation requirements explicitly, including algorithms, metrics, artifacts and evaluation protocols. However, many decisions needed to build a faithful repository are implicit in framework defaults, cited implementations and community conventions. Existing paper-to-code agents mainly condition generation on the paper text and transient planning context, without a persistent object that carries paper-derived obligations or binds implicit details to external evidence. As a result, explicit requirements drift out of context across long trajectories, while implicit details fall back on model priors rather than grounded evidence. Faithful reproduction therefore needs a persistent implementation contract that preserves paper-derived obligations and attaches external repository evidence to the implementation decisions it supports.

\system{} instantiates this idea with a global implementation contract containing two channels.
The implementation-requirement channel extracts explicit paper knowledge as code obligations, while the reference-evidence channel grounds implicit repository knowledge in related implementations.
Both channels are bound to work packages and projected into file-level contracts, so the same contract guides generation and serves as the audit target for detecting missing obligations, unsupported implementation choices, broken interfaces, and runtime failures.

\system{} carries this contract through a four-stage pipeline.
Prepare segments the paper, extracts implementation requirements, and builds a reference-repository index.
Plan organizes requirements into work packages, binds package-local reference evidence, derives the target architecture, and freezes file-level contracts over obligations, evidence, validation checks, and file responsibilities.
Generate produces the repository file by file under these local contracts.
Repair audits the result against the same contract together with runtime checks, so failures are reported as violated paper-derived obligations, missing evidence use, broken interfaces, or execution failures rather than only as crashing commands.

We evaluate \system{} on PaperBench Code-Dev~\citep{paperbench2025}, using repository-level rubric grading to test whether generated code implements the target paper rather than merely executing.
We run \system{} with two backbones, Claude-Sonnet-4.5 and Gemini-3-Flash, and under each backbone \system{} reaches the highest mean score among same-backbone scaffolds.
Ablations of the two contract channels together with a per-paper case analysis support that both the implementation-requirement and reference-evidence channels contribute to faithful reproduction.

Our main contributions are:
\begin{itemize}[leftmargin=*,nosep]
    \item We formulate paper-to-code reproduction as contract-guided repository construction under partial paper specifications, where faithful generation requires preserving explicit paper obligations and grounding implicit implementation choices.
    \item We introduce a two-channel implementation contract over paper-derived requirements and related-repository evidence, with work-package bindings, file-level obligations, provenance records, and typed requirement-review/runtime reports that preserve both traces across generation and repair.
    \item We instantiate this contract in \system{}, a four-stage Prepare--Plan--Generate--Repair pipeline, and evaluate it on the full 20-paper PaperBench Code-Dev suite with same-backbone comparisons and ablations that remove each contract channel.
\end{itemize}

\section{Related Work}
\label{sec:related_work}

\paragraph{LLM agents for research and code.}
LLM agents have been applied to research workflows~\citep{researchagent2024,aiscientist2024,agentlab2025}, ML engineering~\citep{mlebench2024,mlbench2025}, and code-generation or repository-level coding settings~\citep{chatdev2024,mapcoder2024,hong2024metagpt,swebench2024,agentless2025,autocoderover2024}.
Paper reproduction differs because the target is a scientific procedure described only partially in natural language: the repository must be faithful to paper-specific methods, protocols, and artifacts, not merely pass a generic execution check.

\paragraph{Paper-to-code reproduction.}
Following DLPaper2Code~\citep{dlpaper2code2018}, recent systems extend reproduction to whole repositories: PaperCoder, the system introduced by Seo et al.~\citep{papercoder2026}, stages planning, analysis, and coding; Sci-Reproducer~\citep{scireproducer2025} highlights missing-algorithm bottlenecks; AutoP2C~\citep{autop2c2025} and HiRAS~\citep{hiras2026} push multimodal parsing and hierarchical supervision; AutoReproduce~\citep{autoreproduce2025} retrieves along citation graphs; DeepCode~\citep{deepcode2026} treats reproduction as agentic coding.
To our knowledge, prior systems do not make both paper-derived obligations and retrieved repository evidence a persistent, file-level audit object shared across planning, generation, and repair, which is the traceability layer \system{} targets.

\paragraph{Reference retrieval and contracts.}
Retrieval-augmented code generation depends on retrieval granularity~\citep{densex2024} and pulls evidence from documentation~\citep{docprompting2023,doccgen2024}, repositories~\citep{repocoder2023,cheng-etal-2024-dataflow}, or code graphs~\citep{coderag2025}, but does not tie each snippet to the scientific requirement it supports.
Our contract view connects to requirements traceability and design by contract~\citep{traceability1994,meyer1992,clelandhuang2014}, adapted to settings where requirements are extracted from paper snippets rather than written formally.
\system{} therefore splits reference evidence into content evidence for reuse or adaptation and structure evidence for architecture planning, and binds both to work-package and file-level contracts that persist across the pipeline.

\paragraph{Evaluation and fidelity.}
PaperBench~\citep{paperbench2025} grades reproductions against rubrics, with related benchmarks studying reproducibility and paper-code mismatch~\citep{yan-etal-2025-lmr,corebench2024,baumgartner2026scicoqa}.
RePro~\citep{repro2025} applies paper-derived criteria as post-hoc refinement, after a candidate repository exists.
\system{} differs by placing the same paper-derived requirements into the forward generation contract, so the audit target is fixed before generation rather than reconstructed afterwards.

\begin{figure*}[t]
  \centering
  \includegraphics[width=0.99\textwidth]{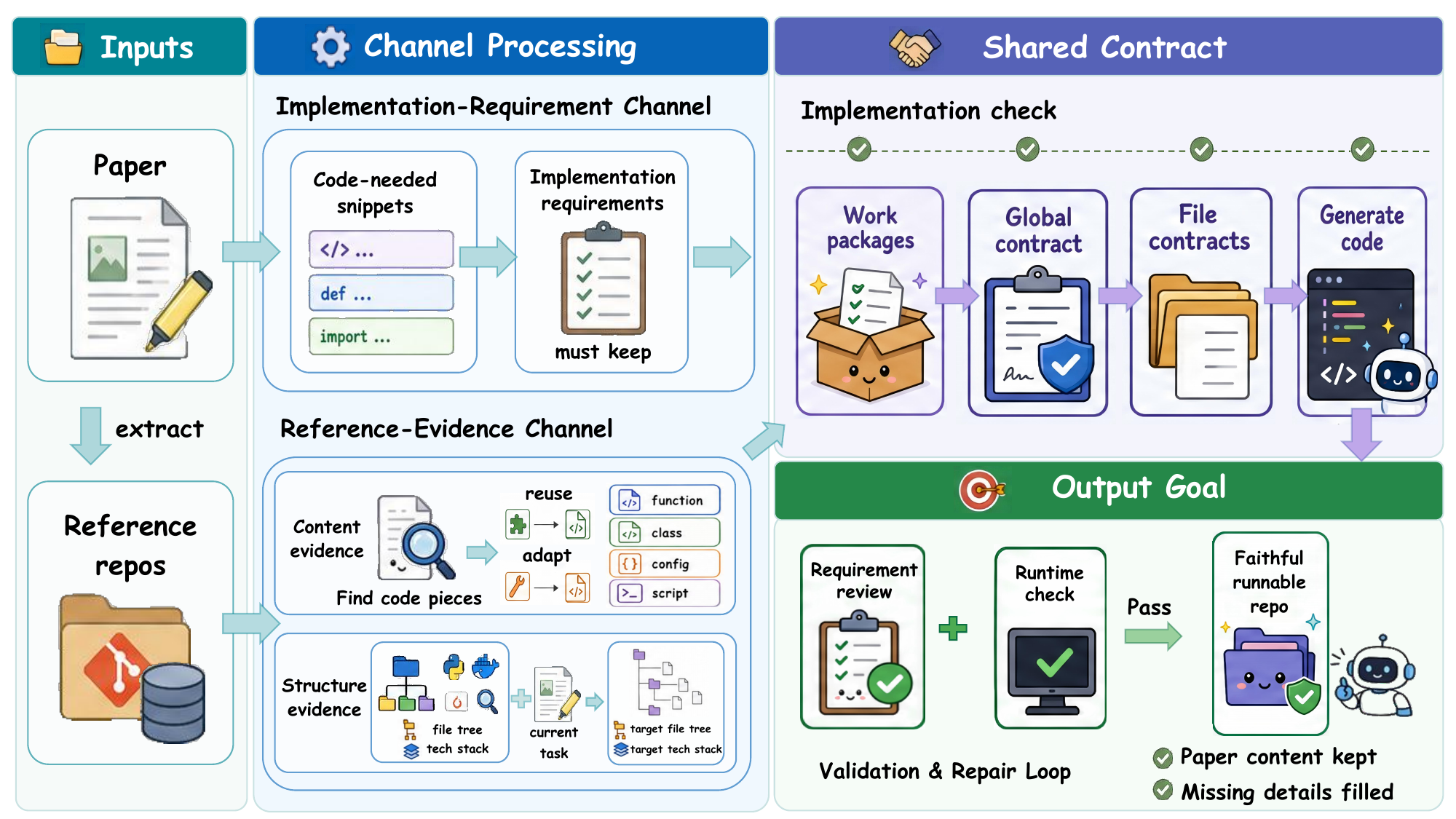}
    \caption{\system{} overview. The implementation-requirement channel extracts code-facing obligations from the target paper, while the reference-evidence channel retrieves content and structure evidence from related repositories. The shared contract binds both channels into work packages, global and file-level contracts, generation, and validation, yielding a repository that is checked for both requirement fidelity and runtime correctness. Representative unit and evidence records appear in Appendix~\ref{sec:appendix_worked_examples}.}
  \label{fig:method_contract}
\end{figure*}

\section{Method}
\label{sec:method}

\system{} turns a paper into a repository by maintaining two persistent contract channels, shown in Figure~\ref{fig:method_contract}. The implementation-requirement channel preserves what the target paper explicitly says must be implemented. The reference-evidence channel supplies evidence for implicit repository knowledge: implementation conventions, file structure, entry points, and protocol details from cited or related repositories. Both channels are bound to work packages and projected into file-level contracts, so information extracted early remains visible during planning, generation, validation, and repair. A work package is a small planning unit that bundles related requirements with the evidence supporting them; we give its full schema in \S\ref{sec:method:work_packages}.

\subsection{Implementation Requirements}
\label{sec:method:reqs}

An implementation requirement is a paper snippet that must be realized in code, and we represent it as a unit
\[
u = (\mathrm{id}, \mathrm{src}, \mathrm{stmt}, \mathrm{cite}, \mathrm{surf}, \mathrm{art}, \mathrm{status}).
\]
Here $\mathrm{id}$ is a stable identifier, $\mathrm{src}$ records the paper source span the unit is drawn from, and $\mathrm{stmt}$ is a normalized statement of the obligation.
The remaining four fields specify how the unit travels through the rest of the pipeline: $\mathrm{cite}$ stores explicit reference cues when they are present in the paper, and the reference-evidence channel attaches prior-work snippets through unit IDs, source anchors, and match records (\S\ref{sec:method:reference_evidence}); $\mathrm{surf}$ names the expected code surfaces, such as a model class, loss function, optimizer, metric, CLI option, config field, training script, or artifact writer; $\mathrm{art}$ enumerates the artifacts the unit should produce; and $\mathrm{status}$ is a lifecycle flag that stays active until the unit is realized.
A concrete example of such a unit is given in Appendix~\ref{sec:appendix_unit_example}.

\paragraph{Extraction.}
\label{sec:method:extract}
We first segment the paper into section-level chunks, falling back to paragraph-level splits when a section is too long for a single LLM pass.
From each chunk, we extract only snippets that imply concrete implementation work, such as algorithmic steps, loss formulations, optimizer and training settings, dataset and preprocessing rules, and reported metrics or artifacts; pure motivation, related work narration, and discussion are skipped.

\paragraph{Preservation invariant.}
To make this preservation operational, we attach a coverage invariant to every active unit so that any silent loss surfaces as a checkable failure inside the pipeline.
For every active unit $u \in \mathcal{U}_{\mathrm{active}}$, the contract requires
\begin{equation*}
\label{eq:invariant}
\begin{aligned}
&\operatorname{own}(u),\, \operatorname{proj}(u),\, \operatorname{prov}(u) \neq \emptyset, \\
&\operatorname{verdict}(u) \in \mathcal{Y},
\end{aligned}
\end{equation*}
where $\mathcal{Y}$ is the fixed verdict set returned by requirement review, with four possible values: pass, missing, wrong-surface and inconsistent-with-source.
The four predicates correspond to four pipeline stages:

\begin{itemize}[leftmargin=*,nosep]
\item \textbf{Ownership.} Plan assigns the unit to at least one work package.
\item \textbf{File projection.} File planning assigns the unit to at least one file contract whose declared surfaces cover the unit's expected surfaces.
\item \textbf{Provenance.} Generation emits a record linking the produced file region back to the unit.
\item \textbf{Verdict.} Requirement review returns a value in $\mathcal{Y}$.
\end{itemize}
A failed predicate creates a contract violation report and a repair obligation; if the repair budget is exhausted, the unresolved verdict is retained in the diagnostics rather than silently absorbed.

\subsection{Reference Evidence}
\label{sec:method:reference_evidence}

Papers often leave implementation details implicit: a paper may cite a prior method without restating its update rule, or refer to a benchmark's ``standard'' data augmentation whose canonical form lives in a reference repository.
\system{} therefore collects reference evidence from related repositories listed in the paper's bibliography, and represents each evidence item as
\[
e = (\mathrm{repo}, \mathrm{path}, \mathrm{region}, \mathrm{units}, \mathrm{role}, \mathrm{basis}).
\]
Here $\mathrm{repo}$ identifies the source repository, $\mathrm{path}$ the file inside it, and $\mathrm{region}$ the matched line range; $\mathrm{units}$ lists the requirement IDs the snippet supports, $\mathrm{role}$ records whether the snippet is meant for local reuse or for adaptation as an interface pattern, and $\mathrm{basis}$ is a structured match record explaining why the snippet is relevant.
Each item is bound to the work package that owns its supported units, so reference code enters generation through a specific package contract rather than as global context.
A concrete evidence item is given in Appendix~\ref{sec:appendix_evidence_example}.

\paragraph{Content evidence.}
For each candidate reference repository, \system{} indexes file paths, imports, exported symbols, scripts, configurations, and entry points described in the README.
A snippet is retained when one of three signals connects it to an owned unit, an expected artifact, or a declared implementation surface: a symbol or interface name match, a path or configuration cue, or a structured LLM judgment.
The judgment is stored in $\mathrm{basis}$ as the supported unit IDs, the matching surface or artifact, and whether the snippet should be reused locally or adapted as an interface pattern, rather than as a free-form note.
The $\mathrm{role}$ field then acts as a use constraint: local reuse covers reusable formulas or transforms, while interface adaptation covers structural patterns that should be re-implemented under the package's own interface, not copied wholesale.

\paragraph{Structure evidence.}
Reference repositories also expose where code should live: file trees, entry scripts, module boundaries, configuration conventions, dependency files, and technology-stack hints.
Architecture planning combines these structural priors with the task and requirements to decide the target file tree, stable interfaces, entry points, and dependency graph.
Content evidence therefore informs how to implement an obligation, while structure evidence informs where the implementation should live.

\subsection{Work Packages and the Global Contract}
\label{sec:method:work_packages}

A work package is the local planning unit that binds requirements to evidence.
Each package records its goal, the requirement IDs it owns, its dependencies on other packages, the public interface it exposes, the artifacts it should produce, and the evidence items attached to it.

After evidence binding, the planner freezes a global contract
\begin{equation*}
\mathcal{G} = (\mathcal{U},\, \mathcal{W},\, \mathcal{E},\, \mathcal{A},\, \mathcal{V}),
\end{equation*}
where $\mathcal{U}$ is the active requirement set, $\mathcal{W}$ the work packages, $\mathcal{E}$ the package-bound evidence, $\mathcal{A}$ the target architecture, and $\mathcal{V}$ the validation plan.
Once $\mathcal{G}$ is frozen, later stages can refine local implementation choices and add derived helper files, but the active requirements, package ownerships, and evidence-package bindings carried by $\mathcal{G}$ stay fixed.
File planning then projects $\mathcal{G}$ into file contracts $\mathcal{C}_F = \{c_f\}$, where each $c_f$ specifies its target path, owner package, owned units, imported and exported symbols, attached evidence, produced artifacts, and review checks.
Every generated file therefore has a per-file specification grounded in the global contract.

\subsection{Generation, Validation, and Repair}
\label{sec:method:repair}

Generation proceeds file by file in dependency order.
Each call is conditioned on the file contract $c_f$, the evidence attached to its owner package, and the already-generated dependency files needed for the local interface; reference code from sibling packages is not injected as global context.
This package-local scoping keeps each produced file auditable: a training loop is generated against training requirements and training evidence, while an evaluation script is generated against metric and artifact requirements.
Each generated file returns a provenance record linking file regions back to the requirement and evidence IDs they realize, and lists implementation choices that are inferred rather than directly specified by the contract.

Validation produces two typed reports.
Requirement review evaluates each active unit against its file contract and provenance record, and returns a verdict in $\mathcal{Y}$.
Runtime validation runs dependency installation, import checks, declared entry points, smoke commands, and benchmark commands when available.
The two reports surface different failures: runtime validation catches broken code and missing artifacts, while requirement review catches runnable but paper-inconsistent implementations.

When either report fails, \system{} issues a repair ticket rather than regenerating the repository.
The ticket carries the failed units, target files, attached evidence, the edits required, and the units that already passed and must remain protected.
Repair is file-local unless the ticket identifies an interface-level failure that requires coordinating two or more files, and repository regions whose contracts are unchanged skip redundant validation.
The repair loop is bounded by a fixed budget; we use the same budget within each backbone setting and report the setting-specific value in \S\ref{sec:implementation}.

\section{Experimental Setup}
\label{sec:experiments}

\subsection{Benchmark and Backbones}
\label{sec:benchmark}

We evaluate on \textbf{PaperBench Code-Dev}~\citep{paperbench2025}, which contains 20 ICML 2024 papers and hierarchical rubrics for repository-level code reproduction; the rubrics grade whether a generated repository implements paper-specific methods, protocols, and artifacts rather than only running.
On this benchmark, the evaluation has three layers.
The full-suite layer compares the primary \system{} run with reported PaperBench Code-Dev systems, using Claude-Sonnet-4.5\footnote{\url{https://platform.claude.com/docs/en/about-claude/models/model-ids-and-versions}.} as the generation backbone.
The same-backbone layer compares \system{} with simpler scaffolds under Gemini-3-Flash\footnote{\url{https://ai.google.dev/gemini-api/docs/gemini-3}.}, the backbone shared with BasicAgent, IterAgent and AiScientist.
The ablation layer removes each contract channel under the same Gemini-3-Flash backbone.

\subsection{Comparisons and Metrics}
\label{sec:systems}

External rows reuse the model, scaffold, and runtime budget reported by their source.
The AiScientist engineering report provides the BasicAgent, IterAgent, and AiScientist scaffold rows used in the same-backbone comparison.
The aggregate full-suite figure includes only Code-Dev rows, and the per-paper table expands the main comparison to BasicAgent, IterAgent, AiScientist, DeepCode, and both \system{} backbones; Appendix~\ref{sec:external_baselines} lists the additional external rows.

We follow the Code-Dev judging protocol and report macro-averaged PaperBench score as the primary metric: weighted leaf scores aggregate to a per-paper score, and the 20 paper scores are averaged.
Token, time, cost, and judge-provenance diagnostics are reported in the appendix.

\subsection{Implementation and Ablations}
\label{sec:implementation}

\system{} is implemented as a four-stage pipeline: the paper is chunked by section, reference repositories come from PaperBench metadata and related-code search subject to the benchmark blacklist on official target repositories, and planning freezes a global contract before file-level generation.
Repair combines requirement review with sandboxed runtime validation, so failures become targeted tickets that patch only affected files.
We cap stage-review repair at three attempts and repository-level repair at five rounds, uniformly across the Claude-Sonnet-4.5 main run and all Gemini-3-Flash full and ablation runs; holding this budget fixed isolates the channel effect from repair-budget effects.
In total we report 20 papers $\times$ 4 \system{} settings = 80 reproduction runs, one run per (paper, setting) pair.

\paragraph{Channel ablations.}
\label{sec:ablation_conditions}
On the same Gemini-3-Flash backbone, No Implementation Requirements skips requirement extraction and requirement review, so architecture planning falls back on the paper's section outline and reference-derived module structure; No Reference Evidence skips reference-repository surveying, so generation relies on requirement-derived contracts without bound content or structure evidence.
All other components and repair budgets are held fixed.

\section{Results and Analysis}
\label{sec:results}

\begin{figure}[t]
  \centering
  \includegraphics[width=\columnwidth]{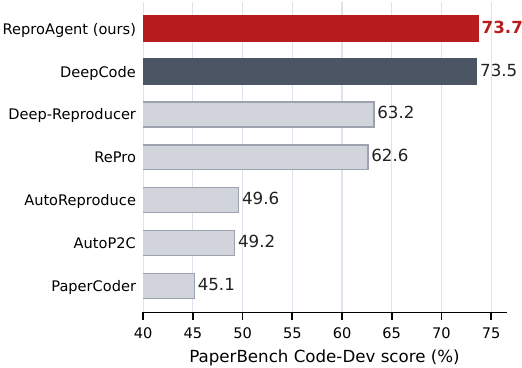}
  \caption{Full-suite PaperBench Code-Dev comparison across reported paper-to-code systems.}
  \label{fig:paperbench_main_comparison}
\end{figure}

\begin{table*}[t]
\centering
\tbcompact
\setlength{\tabcolsep}{3pt}
\renewcommand{\arraystretch}{1.05}
\begin{tabularx}{0.96\textwidth}{@{}L{0.36\textwidth}YYYC{0.16\textwidth}@{}}
\toprule
\textbf{Paper} & \textbf{BasicAgent} & \textbf{IterAgent} & \textbf{DeepCode}
& \multicolumn{1}{c}{\textbf{\mbox{\system{} (ours)}}} \\
\midrule
Adaptive Pruning & 16.6\tbzero{0.0} & 22.4\tbpos{5.8} & \tbsecond{54.4\tbpos{37.8}} & \tbbest{67.8\tbpos{51.2}} \\
All in One & 24.8\tbzero{0.0} & 11.5\tbneg{13.3} & \tbsecond{75.9\tbpos{51.1}} & \tbbest{76.9\tbpos{52.1}} \\
BAM & 25.2\tbzero{0.0} & 18.7\tbneg{6.5} & \tbbest{74.8\tbpos{49.6}} & \tbsecond{61.8\tbpos{36.6}} \\
BBOX & 10.7\tbzero{0.0} & 15.2\tbpos{4.5} & \tbsecond{64.4\tbpos{53.7}} & \tbbest{86.4\tbpos{75.7}} \\
Bridging Data Gaps & 15.8\tbzero{0.0} & 8.1\tbneg{7.7} & \tbsecond{58.1\tbpos{42.3}} & \tbbest{66.0\tbpos{50.2}} \\
FRE & 22.7\tbzero{0.0} & 17.5\tbneg{5.2} & \tbbest{81.4\tbpos{58.7}} & \tbsecond{76.8\tbpos{54.1}} \\
FTRL & 9.3\tbzero{0.0} & 4.3\tbneg{5.0} & \tbsecond{59.8\tbpos{50.5}} & \tbbest{63.9\tbpos{54.6}} \\
LBCS & 19.8\tbzero{0.0} & 20.7\tbpos{0.9} & \tbbest{74.7\tbpos{54.9}} & \tbsecond{65.8\tbpos{46.0}} \\
LCA on the Line & 14.0\tbzero{0.0} & 13.3\tbneg{0.7} & \tbbest{74.9\tbpos{60.9}} & \tbsecond{64.9\tbpos{50.9}} \\
Mechanistic Understanding & 18.1\tbzero{0.0} & 3.7\tbneg{14.4} & \tbbest{92.5\tbpos{74.4}} & \tbsecond{75.7\tbpos{57.6}} \\
PINN & 24.8\tbzero{0.0} & 22.5\tbneg{2.3} & \tbbest{91.0\tbpos{66.2}} & \tbsecond{81.5\tbpos{56.7}} \\
RICE & 19.8\tbzero{0.0} & 7.8\tbneg{12.0} & \tbsecond{70.2\tbpos{50.4}} & \tbbest{75.1\tbpos{55.3}} \\
Robust CLIP & 29.6\tbzero{0.0} & 24.7\tbneg{4.9} & \tbbest{73.3\tbpos{43.7}} & \tbsecond{63.0\tbpos{33.4}} \\
Sample-specific Masks & 30.9\tbzero{0.0} & 25.7\tbneg{5.2} & \tbsecond{67.1\tbpos{36.2}} & \tbbest{82.0\tbpos{51.1}} \\
SAPG & 5.4\tbzero{0.0} & 6.3\tbpos{0.9} & \tbsecond{73.8\tbpos{68.4}} & \tbbest{86.5\tbpos{81.1}} \\
Sequential Neural Score Estimation & 41.4\tbzero{0.0} & 32.7\tbneg{8.7} & \tbbest{87.0\tbpos{45.6}} & \tbsecond{78.0\tbpos{36.6}} \\
Stay on Topic with Classifier-Free Guidance & 10.8\tbzero{0.0} & 16.2\tbpos{5.4} & \tbbest{70.5\tbpos{59.7}} & \tbsecond{58.5\tbpos{47.7}} \\
Stochastic Interpolants & 12.4\tbzero{0.0} & 26.3\tbpos{13.9} & \tbbest{81.5\tbpos{69.1}} & \tbsecond{80.8\tbpos{68.4}} \\
Test-Time Model Adaptation & 14.3\tbzero{0.0} & 14.2\tbneg{0.1} & \tbsecond{64.9\tbpos{50.6}} & \tbbest{75.6\tbpos{61.3}} \\
What Will My Model Forget & 29.7\tbzero{0.0} & 11.4\tbneg{18.3} & \tbsecond{80.8\tbpos{51.1}} & \tbbest{87.6\tbpos{57.9}} \\
\midrule
\textbf{Mean score} & 19.8 & 16.2\tbneg{3.6} & \tbsecond{73.5\tbpos{53.7}} & \tbbest{73.7\tbpos{53.9}} \\
\bottomrule
\end{tabularx}
\caption{\textbf{Per-paper PaperBench Code-Dev comparison with Claude backbones.} Claude columns for BasicAgent and IterAgent follow the PaperBench report~\citep{paperbench2025}; DeepCode follows its concurrent report~\citep{deepcode2026}. BasicAgent and IterAgent use Claude-3.5-Sonnet, while DeepCode and \system{} (ours) use Claude-4.5-Sonnet. Subscripts show absolute-point differences from BasicAgent on the same paper; red upward and green downward arrows indicate the sign of the difference. \bestlegend{} and \secondlegend{} mark the row-best and row-second-best scores among the shown Claude columns.}
\label{tab:claude_test_matrix}
\end{table*}

\begin{table*}[t]
\centering
\tbcompact
\setlength{\tabcolsep}{3pt}
\renewcommand{\arraystretch}{1.05}
\begin{tabularx}{0.96\textwidth}{@{}L{0.36\textwidth}YYYC{0.16\textwidth}@{}}
\toprule
\textbf{Paper} & \textbf{BasicAgent} & \textbf{IterAgent} & \textbf{AiScientist}
& \multicolumn{1}{c}{\textbf{\mbox{\system{} (ours)}}} \\
\midrule
Adaptive Pruning & 24.5\tbzero{0.0} & 3.0\tbneg{21.5} & \tbsecond{27.2\tbpos{2.7}} & \tbbest{41.9\tbpos{17.4}} \\
All in One & 20.9\tbzero{0.0} & 45.1\tbpos{24.2} & \tbsecond{46.3\tbpos{25.4}} & \tbbest{51.6\tbpos{30.7}} \\
BAM & \tbsecond{48.5\tbzero{0.0}} & 45.0\tbneg{3.5} & \tbbest{56.6\tbpos{8.1}} & \tbsecond{48.5\tbzero{0.0}} \\
BBOX & 15.4\tbzero{0.0} & 8.3\tbneg{7.1} & \tbbest{33.8\tbpos{18.4}} & \tbsecond{15.6\tbpos{0.2}} \\
Bridging Data Gaps & 12.6\tbzero{0.0} & 12.4\tbneg{0.2} & \tbsecond{23.1\tbpos{10.5}} & \tbbest{50.6\tbpos{38.0}} \\
FRE & 21.7\tbzero{0.0} & 23.9\tbpos{2.2} & \tbbest{35.2\tbpos{13.5}} & \tbsecond{26.9\tbpos{5.2}} \\
FTRL & 5.9\tbzero{0.0} & 4.2\tbneg{1.7} & \tbsecond{10.1\tbpos{4.2}} & \tbbest{40.1\tbpos{34.2}} \\
LBCS & 17.8\tbzero{0.0} & 15.3\tbneg{2.5} & \tbsecond{27.9\tbpos{10.1}} & \tbbest{42.5\tbpos{24.7}} \\
LCA on the Line & 13.0\tbzero{0.0} & 18.3\tbpos{5.3} & \tbsecond{30.2\tbpos{17.2}} & \tbbest{41.8\tbpos{28.8}} \\
Mechanistic Understanding & 14.9\tbzero{0.0} & 21.9\tbpos{7.0} & \tbsecond{29.9\tbpos{15.0}} & \tbbest{48.1\tbpos{33.2}} \\
PINN & 26.6\tbzero{0.0} & 30.8\tbpos{4.2} & \tbsecond{49.9\tbpos{23.3}} & \tbbest{61.3\tbpos{34.7}} \\
RICE & 10.4\tbzero{0.0} & 8.9\tbneg{1.5} & \tbsecond{10.9\tbpos{0.5}} & \tbbest{31.0\tbpos{20.6}} \\
Robust CLIP & 15.4\tbzero{0.0} & 10.4\tbneg{5.0} & \tbsecond{18.3\tbpos{2.9}} & \tbbest{29.5\tbpos{14.1}} \\
Sample-specific Masks & 25.4\tbzero{0.0} & 33.3\tbpos{7.9} & \tbsecond{36.8\tbpos{11.4}} & \tbbest{53.8\tbpos{28.4}} \\
SAPG & 11.4\tbzero{0.0} & 12.7\tbpos{1.3} & \tbsecond{19.9\tbpos{8.5}} & \tbbest{29.0\tbpos{17.6}} \\
Sequential Neural Score Estimation & 53.5\tbzero{0.0} & \tbsecond{60.2\tbpos{6.7}} & \tbbest{64.9\tbpos{11.4}} & 45.7\tbneg{7.8} \\
Stay on Topic with Classifier-Free Guidance & 8.4\tbzero{0.0} & \tbsecond{13.7\tbpos{5.3}} & 20.1\tbpos{11.7} & \tbbest{25.3\tbpos{16.9}} \\
Stochastic Interpolants & 17.0\tbzero{0.0} & 17.4\tbpos{0.4} & \tbsecond{18.8\tbpos{1.8}} & \tbbest{39.1\tbpos{22.1}} \\
Test-Time Model Adaptation & 15.3\tbzero{0.0} & 18.1\tbpos{2.8} & \tbsecond{32.5\tbpos{17.2}} & \tbbest{54.9\tbpos{39.6}} \\
What Will My Model Forget & 6.6\tbzero{0.0} & 9.0\tbpos{2.4} & \tbbest{17.9\tbpos{11.3}} & \tbsecond{16.8\tbpos{10.2}} \\
\midrule
\textbf{Mean score} & 19.3 & 20.6\tbpos{1.3} & \tbsecond{30.5\tbpos{11.2}} & \tbbest{39.7\tbpos{20.4}} \\
\bottomrule
\end{tabularx}
\caption{\textbf{Per-paper PaperBench Code-Dev comparison with Gemini.} Gemini columns for BasicAgent, IterAgent, and AiScientist follow the AiScientist engineering report~\citep{aiscientistengineering2026}; all Gemini columns use Gemini-3-Flash. Subscripts show absolute-point differences from BasicAgent on the same paper; red upward and green downward arrows indicate the sign of the difference. \bestlegend{} and \secondlegend{} mark the row-best and row-second-best scores among the shown Gemini columns.}
\label{tab:gemini_test_matrix}
\end{table*}

\subsection{Full-Suite Comparison}
\label{sec:main_results}

Figure~\ref{fig:paperbench_main_comparison} reports mean PaperBench Code-Dev scores across published paper-to-code systems.
\system{} with Claude-Sonnet-4.5 reaches \reproagentScore{}, above PaperCoder~\citep{papercoder2026} (45.1), AutoP2C~\citep{autop2c2025} (49.2), AutoReproduce~\citep{autoreproduce2025} (49.6), and Deep-Reproducer~\citep{deepreproducer2025} (63.2), and edges past the concurrent state-of-the-art DeepCode at \deepcodeScore{}.

Tables~\ref{tab:claude_test_matrix} and~\ref{tab:gemini_test_matrix} expand this view into the 20 targets for each backbone; additional external rows appear in Appendix~\ref{sec:external_baselines}.
We hypothesize that the gain comes from the contract itself: requirement and evidence bindings that survive across files keep dependent artifacts, such as method, training, and evaluation code, consistent during generation and repair instead of being re-derived per stage.

\subsection{Same-Backbone Comparison}
\label{sec:controlled_comparison}

Holding the backbone fixed at Gemini-3-Flash, \system{} reaches \geminiFullScore{} against \basicAgentScore{} for BasicAgent, \iterAgentScore{} for IterAgent, and \aiScientistScore{} for AiScientist.
The advantage over the strongest external Gemini row is on the order of 9 absolute points, consistent with the contract effect we hypothesize in the full-suite comparison.

\subsection{Ablation Study}
\label{sec:ablation}

\begin{figure}[t]
  \centering
  \includegraphics[width=0.93\columnwidth]{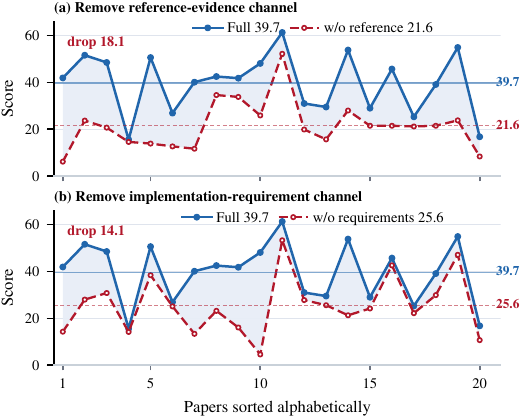}
  \caption{Gemini channel ablations on all 20 papers. Removing reference evidence (top) or implementation requirements (bottom) lowers the mean; judge provenance and token diagnostics are in the appendix.}
  \label{fig:gemini_ablation_stacked}
\end{figure}

\begin{table}[t]
\centering
\small
\setlength{\tabcolsep}{3pt}
\begin{tabular}{@{}lccc@{}}
\toprule
Setting & Mean & Median & $\Delta$ \\
\midrule
Full contract & \geminiFullScore{} & \geminiFullMedianScore{} & --- \\
\quad $-$ Reference evidence & \geminiNoRefScore{} & \geminiNoRefMedianScore{} & $-$\geminiNoRefDrop{} \\
\quad $-$ Implementation requirements & \geminiNoReqScore{} & \geminiNoReqMedianScore{} & $-$\geminiNoReqDrop{} \\
\bottomrule
\end{tabular}
\caption{Gemini-3-Flash ablation results over 20 papers; $\Delta$ is the mean drop from the full contract.}
\label{tab:gemini_ablation}
\end{table}

We next isolate the contribution of each contract channel under Gemini-3-Flash.
Table~\ref{tab:gemini_ablation} summarizes the aggregates: removing reference evidence drops the mean by \geminiNoRefDrop{} and removing implementation requirements drops it by \geminiNoReqDrop{}.
The full setting beats both ablations on every one of the 20 targets, so each channel contributes capability that the other cannot recover.

\paragraph{Requirement channel: Sample-specific Masks.}
Sample-specific Masks scores 53.8 with the full contract, 21.3 without implementation requirements, and 28.0 without reference evidence: the requirement drop is the dominant one.
The paper's contribution is a small set of named, implementation-critical obligations, including a per-sample mask generator, interpolation from a low-resolution mask up to the input image size, joint optimization of a shared pattern and the mask-generator parameters, the PAD/NARROW/MEDIUM/FULL baselines, and ablations over shared-pattern and mask channels.
The full contract pins these obligations to specific work packages and files, so the generated repository keeps the baseline and ablation method registries that the rubric checks.
Without the requirement channel, the planner falls back on the section outline and reference-derived modules, and the generated repository keeps runnable scaffolding but drops the named baselines, ablations, and mask-channel switches, since none of them are pinned to a file.
The case-level code excerpt and a per-paper score table for additional cases appear in Appendix~\ref{sec:case_study_ttma}.

\paragraph{Evidence channel: Bridging Data Gaps.}
Bridging Data Gaps scores 50.6 with the full contract, 38.4 without implementation requirements, and 13.9 without reference evidence: the evidence drop is the dominant one.
The paper names the high-level task and adaptor objectives, but its concrete protocol surface, including file layout, training-loop wiring, checkpoint and trace artifacts, and configuration handling, is supplied by reference repositories rather than the paper text.
With the full contract, the generated repository emits the expected checkpoint, training-trace, and resolved-configuration artifacts under conventional output paths, matching the pattern of checkpointed training and reproducible configuration logging.
Without the evidence channel, paper obligations remain visible in the plan, but the generated repository under-implements file layout, data-handling conventions, experiment wiring, and artifact naming, so it fails to produce these expected artifacts.
Appendix~\ref{sec:case_study_ttma} gives the corresponding code-level trace and Appendix Table~\ref{tab:appendix_our_scores_files} lists the per-paper scores behind these aggregates.

\subsection{Discussion}
\label{sec:result_interpretation}

Two observations stand out.
First, the same-backbone advantage and the full-suite gain move together, so the lift attaches to the contract rather than to any single backbone.
Second, the two channels are complementary: removing either drops the mean by more than 14 points, and the failures fall into two classes, paper-specific obligations dropped without the requirement channel and repository-level scaffolding lost without the evidence channel.
Together, these support a mechanistic claim: \system{} benefits from a global contract whose requirement channel preserves explicit paper obligations and whose evidence channel grounds implicit repository knowledge.

\section{Conclusion}
\label{sec:conclusion}

We presented \system{}, a contract-guided pipeline for paper-to-code reproduction.
Its global contract binds explicit paper obligations and implicit reference-repository evidence to work packages, file contracts, and repair decisions.
On PaperBench Code-Dev, \system{} scores \reproagentScore{} across all 20 papers; same-backbone rows support scaffold-level gains, while channel ablations and case traces support the role of both contract channels.
The takeaway is that paper obligations and reference-repository evidence are most useful when persisted as a contract: once both are bound, the same artifact anchors planning, exposes coverage failures, and scopes repair.

\section{Limitations}
\label{sec:limitations}

\system{} targets repository-level reproduction in which the paper and benchmark-allowed reference material jointly specify the implementation target.
Extending the approach to other reproduction settings, modalities, and evaluation protocols is a natural direction for future work.

We report token, time, and call diagnostics in Appendix~\ref{sec:usage_diagnostics} to make the process auditable rather than to compare deployment cost; further engineering optimizations are compatible with the contract design and are also left to future work.

\bibliography{refs}

\clearpage
\appendix

\section{Contract Case Study: Unit and Evidence Records}
\label{sec:appendix_worked_examples}

This appendix gives two paired schema records for a continual-RL paper that introduces a forgetting-aware fine-tuning protocol.
The first illustrates the implementation-requirement schema in \S\ref{sec:method:reqs}; the second illustrates the reference-evidence schema in \S\ref{sec:method:reference_evidence} and is anchored to the unit from the first.
The records are compact schema views that preserve source anchors and evidence bindings used by review and repair.

\subsection{An Implementation-Requirement Unit}
\label{sec:appendix_unit_example}

The unit below illustrates how Prepare extracts a requirement, Plan assigns it to a method work package, Generate projects it into a model-method file contract, and Repair verifies it with requirement review.

\begin{repocode}{Schema-aligned \texttt{units.json} view for the EWC running example}
{
  "unit_id": "unit_004",
  "type": "method",
  "statement": "implement Elastic Weight Consolidation (EWC)",
  "paper_evidence": [
    "L_EWC = L_RL + (lambda/2) * sum_i F_i (theta_i - theta_*_i)^2",
    "where theta_* is the pre-trained model and F is the diagonal of the Fisher matrix."
  ],
  "source_paragraph_ids": ["chunk_004_02", "chunk_024"],
  "verification_targets": [
    {
      "kind": "artifact",
      "description": "Fisher matrix calculation and EWC loss implementation"
    }
  ],
  "implementation_surfaces": ["model_or_method"],
  "code_obligations": [
    "Implement diagonal Fisher Information Matrix from the pre-trained policy on its training distribution.",
    "Implement the EWC loss term (lambda/2) * sum_i F_i (theta_i - theta_*_i)^2.",
    "Use lambda=10000 as the default coefficient."
  ],
  "status": "active"
}
\end{repocode}

\paragraph{What each field stores.}
\begin{itemize}[leftmargin=*,nosep]
  \item \textbf{unit\_id} is a stable identifier; repair tickets and provenance records both index by this value, so the same unit remains addressable across stages.
  \item \textbf{type} categorizes the unit (here: method) and drives how Plan groups it with related units into a work package.
  \item \textbf{statement} is the normalized one-duty implementation requirement, derived from the paper rather than copied verbatim.
  \item \textbf{paper\_evidence} keeps the originating paper quotes, so review can cross-check the statement against the source.
  \item \textbf{source\_paragraph\_ids} are anchors into the segmented paper (chunk IDs), used to locate the unit in the addendum or main body and to guide review and repair.
  \item \textbf{verification\_targets} declares what reviewers should check; each target has a kind (artifact, surface, or metric) and a description.
  \item \textbf{implementation\_surfaces} declares where the obligation should appear in code; file planning must project the unit onto at least one file whose declared surfaces cover this set.
  \item \textbf{code\_obligations} expand the statement into concrete sub-duties that generation and review can verify pointwise.
  \item \textbf{status} is the lifecycle flag; review writes verdicts (pass, missing, wrong-surface, inconsistent-with-source) back into this field.
\end{itemize}

\paragraph{How the preservation invariant fires on this unit.}
Plan cannot proceed unless the EWC unit is bound to at least one owner work package; in this run it is owned by a method package together with the related PPO-fine-tuning unit.
File planning cannot proceed unless the unit is projected into at least one file contract whose declared surface includes \texttt{model\_or\_method}; in this run the unit lands in a single dedicated method file.
Generate cannot finalize that file unless the file emits a provenance record citing \texttt{unit\_004}, so the unit id surfaces in the per-file provenance map alongside its source paragraphs.
Repair cannot mark the unit \texttt{pass} until requirement review verifies all three code obligations against the file contract; otherwise it issues a repair ticket carrying the unit id, the matched evidence ids attached to the work package, and the protected units that already passed.

The unit is therefore not a free-form paragraph or a planning-time prompt fragment.
It is an addressable record with a fixed schema that planning, file projection, generation, validation, and repair are all conditioned on, which is what allows the same paper requirement to remain visible across the four-stage pipeline rather than being silently dropped.

\subsection{A Reference-Evidence Item}
\label{sec:appendix_evidence_example}

The evidence item below illustrates how a selected reference snippet is scoped to the method work package that owns the EWC unit \texttt{unit\_004}.

\begin{repocode}{Schema-aligned \texttt{evidence\_bundles.json} view tied to unit\_004}
{
  "work_package_id": "wp_methods",
  "unit_id": "unit_004",
  "ref_id": "ref_001",
  "file_path": "model.py",
  "snippet_preview": "model.py:13:     def __init__(self, in_features, out_features, sigma0=0.5):\nmodel.py:14:         super().__init__()\nmodel.py:15:         self.in_features = in_features\nmodel.py:16:         self.out_features = out_features\nmodel.py:17:         self.weight = nn.Parameter(torch.Tensor(out_features, in_features))\nmodel.py:18:         self.bias = nn.Parameter(torch.Tensor(out_features))\nmodel.py:19:         ...",
  "why_relevant": "function `__init__` in `model.py` aligns with units paper_method_core, unit_004 via parameter construction patterns reusable for the EWC reference network.",
  "confidence": 1.0,
  "matched_keywords": ["nn.Parameter", "in_features", "out_features"]
}
\end{repocode}

\paragraph{What each field stores.}
\begin{itemize}[leftmargin=*,nosep]
  \item \textbf{work\_package\_id} names the package this evidence is attached to; in \S\ref{sec:method:reference_evidence} the same identifier serves as the binding point between requirements and evidence.
  \item \textbf{unit\_id} names the supported requirement; here it is the EWC unit \texttt{unit\_004} from Appendix~\ref{sec:appendix_unit_example}, which is what makes the evidence consumable by the same package contract.
  \item \textbf{ref\_id}, \textbf{file\_path}, and \textbf{snippet\_preview} together pin down $(\mathrm{repo}, \mathrm{path}, \mathrm{region})$: the source repository, the file inside it, and the matched line range, recorded so review and repair can locate the snippet again.
  \item \textbf{why\_relevant} is the structured match basis: it explains, in one sentence, which symbol or interface aligns the snippet with the supported unit, rather than leaving the LLM judgment as a free-form note.
  \item \textbf{confidence} and \textbf{matched\_keywords} record the auxiliary signals used by the retain rule in \S\ref{sec:method:reference_evidence}, namely a symbol-level match strengthened by an LLM judgment.
  \item The role field, omitted from this stripped record but tracked elsewhere in the run state, marks this snippet as reusable structural evidence for parameter construction.
\end{itemize}

\paragraph{How the evidence enters generation.}
The item is bound to \texttt{wp\_methods} during Plan, so when Generate writes the file contract that owns \texttt{unit\_004}, the prompt for that file sees this snippet but no unrelated reference code from sibling work packages.
If requirement review later returns a non-pass verdict on \texttt{unit\_004}, the repair ticket carries the same \texttt{ref\_id} and \texttt{file\_path}, allowing repair to re-anchor the edit on this specific reference snippet rather than searching the reference repository again.

\section{External Baselines and Cost Notes}
\label{sec:external_baselines}

Table~\ref{tab:appendix_external_baselines} records aggregate external rows that are useful context but not part of the primary same-backbone comparison.
Rows include official PaperBench results, recent paper-to-code systems, and separately collected external reference rows.
They differ in scaffold, model, runtime budget, and reporting protocol, and provide contextual reported scores rather than a primary controlled comparison.

\begin{table*}[t]
\centering
\small
\setlength{\tabcolsep}{4pt}
\newcommand{\baselinegroup}[2]{\multirow[c]{#1}{0.40\textwidth}{\centering\arraybackslash #2}}
\begin{tabular*}{\textwidth}{@{\extracolsep{\fill}}>{\centering\arraybackslash}m{0.40\textwidth}>{\centering\arraybackslash}m{0.30\textwidth}>{\centering\arraybackslash}m{0.14\textwidth}@{}}
\toprule
Method & Backbone / setting & Reported score \\
\midrule
\baselinegroup{5}{BasicAgent~\citep{paperbench2025}\textsuperscript{a}} & Gemini-2.0-Flash & 5.0 \\
 & o3-mini & 5.1 \\
 & GPT-4o & 7.7 \\
 & o1 & 19.5 \\
 & Claude-3.5-Sonnet & 35.4 \\
\midrule
\baselinegroup{3}{IterativeAgent~\citep{paperbench2025}\textsuperscript{a}} & o3-mini & 16.4 \\
 & Claude-3.5-Sonnet & 27.5 \\
 & o1 & 43.3 \\
\midrule
PaperCoder~\citep{papercoder2026} & o3-mini-high & 45.1 \\
AutoP2C~\citep{autop2c2025} & source-reported setting & 49.2 \\
AutoReproduce~\citep{autoreproduce2025} & source-reported setting & 49.6 \\
Deep-Reproducer~\citep{deepreproducer2025} & GPT-5 & 63.2 \\
RePro~\citep{repro2025}\textsuperscript{b} & o3-mini & 62.6 \\
DeepCode~\citep{deepcode2026}\textsuperscript{c} & Claude-Sonnet-4.5 & 73.5 \\
\midrule
BasicAgent~\citep{aiscientistengineering2026}\textsuperscript{d} & Gemini-3-Flash & 19.3 \\
IterAgent~\citep{aiscientistengineering2026}\textsuperscript{d} & Gemini-3-Flash & 20.6 \\
AiScientist~\citep{aiscientistengineering2026}\textsuperscript{d} & Gemini-3-Flash & 30.5 \\
\midrule
\baselinegroup{2}{\system{} (ours)} & Claude-Sonnet-4.5 & 73.7 \\
 & Gemini-3-Flash & 39.7 \\
\bottomrule
\end{tabular*}
\caption{Supplemental aggregate PaperBench baselines. Reported scores are percentages. Rows differ in scaffold, backbone, runtime budget, and reporting protocol, and provide context for the full-suite comparison rather than a primary controlled comparison.}
\label{tab:appendix_external_baselines}
\begin{minipage}{0.98\textwidth}
\footnotesize
\textsuperscript{a} Official 20-paper PaperBench Code-Dev scaffold rows; standard errors are reported in the cited source, and this table reports means only for compact contextual comparison. The o1 IterativeAgent row uses two seeds.
\textsuperscript{b} RePro reports PRroot@5, so we list it as external PRroot context separate from the Code-Dev rows.
\textsuperscript{c} Collected 20-paper mean over three runs per paper; the reported mean standard error across papers is 2.8 and the mean cost is \$8.88.
\textsuperscript{d} Collected AiScientist engineering aggregates; mean costs are \$6.25, \$27.44, and \$15.67 for BasicAgent, IterAgent, and AiScientist.
\end{minipage}
\end{table*}

\section{Gemini Ablation Case Studies}
\label{sec:case_study_ttma}

We keep qualitative ablation cases in the appendix so that the main results focus on aggregate behavior.
The examples below illustrate the channel roles behind the aggregate ablation pattern in Section~\ref{sec:ablation}.
Scores in this section provide mechanism-oriented evidence for the aggregate analysis.

\subsection{Tracing One Unit in Test-Time Model Adaptation}
\label{sec:ttma_trace}

We trace Test-Time Model Adaptation because its FOA/CMA unit is a strong trace for the implementation-requirement channel and a useful contrast for how reference evidence affects repository organization.
In our consolidated 2026-05-24 PaperBench judging snapshot, the full Gemini run scores 54.9, while removing implementation requirements scores 47.1 and removing reference evidence scores 23.8.
The trace is taken from the corresponding Gemini full-run artifacts for \texttt{test-time-model-adaptation}.

\paragraph{Extraction.}
\texttt{unit\_002} is extracted as ``Implement Forward-Only Prompt Adaptation with CMA''.
Its obligations require prompt insertion into the ViT input sequence, a CMA-ES optimizer, a population size $K\in[2,28]$, and a forward-only prompt update loop.
\texttt{unit\_003} is extracted as the activation-distribution discrepancy fitness function, requiring mean/variance statistics for test CLS tokens and a fitness value passed to CMA.

\paragraph{Work package and contract.}
Planning binds both units to \texttt{wp\_foa\_core}, whose goal is to implement the FOA method and CMA optimizer.
The work package produces \texttt{methods/foa.py}, \texttt{optimizers/cma\_es.py}, and \texttt{results/source\_stats.json}.
Its global contract preserves the FOA loop, CMA-ES interface, source-statistics collector, fitness function, method registry, and parameter inventory for $K$, prompt length, alignment weights, batch size 64, and momentum 0.9.
This is the implementation-requirement channel in concrete form: the paper obligations become file-owned duties rather than prose context.

\paragraph{Evidence binding.}
The evidence bundle for \texttt{wp\_foa\_core} records paper chunks for FOA, source-statistics calculation, and the evaluation protocols.
In this case, the package is marked \texttt{self\_contained} and has no repository \texttt{evidence\_links}, because the decisive FOA/CMA obligations are already present in the paper and addendum evidence.
This clarifies the ablation: the no-reference setting still recovers much of the requirement surface, while the full setting adds stronger repository-level organization, protocol wiring, and comparison surfaces.

\paragraph{Generated files and provenance.}
Generation materializes the contract in three files.
\texttt{methods/foa.py} contains the hyperparameter sweeps, method/baseline registries, a CMA-ES implementation, source-statistics loading, and FOA routines.
\texttt{optimizers/cma\_es.py} exposes a standalone CMA-ES \texttt{ask}/\texttt{tell} optimizer and method selectors.
\texttt{src/foa/utils/cma\_optimizer.py} contains a forward-only \texttt{FOA.adapt} route under \texttt{torch.no\_grad()}.
The generation file-provenance record assigns all three files to \texttt{wp\_foa\_core}, lists \texttt{unit\_002} and \texttt{unit\_003} as owned units, and attaches validation checks for implementation requirements, formula/algorithm contracts, evidence-matrix coverage, and route closure.
The key point is that the requirement IDs, work-package owner, generated files, and validation checks remain connected throughout the trajectory; the ablation removes one of these channels before the file-level contract is formed.

\subsection{Tracing Reference Evidence in Bridging Data Gaps}
\label{sec:bdg_trace}

Bridging Data Gaps is a complementary case because the same snapshot shows an asymmetric ablation: the full Gemini run scores 50.6, the run without implementation requirements scores 38.4, and the run without reference evidence scores 13.9.
The case therefore stresses the second channel in Figure~\ref{fig:method_contract}: the paper requirement channel keeps the high-level adaptation and artifact-writing obligations visible, while reference evidence helps convert those obligations into a concrete repository protocol.

\paragraph{Requirement channel.}
The requirement units preserve the paper-derived duties that must survive into code: train the adaptation components, expose a runnable experiment path, save model/adaptor checkpoints, and write result artifacts that make the experiment inspectable.
These duties are projected into file contracts that expect a training route and artifact writer rather than a single monolithic script.
This explains the asymmetric score pattern: the remaining reference and generic task context can still support some repository structure, but the paper-specific obligations are less explicit without the requirement channel.

\paragraph{Reference-evidence channel.}
The reference channel supplies the implementation texture that the paper does not spell out line by line.
In the full run, the generated repository writes \texttt{checkpoints/adaptor.pth}, \texttt{checkpoints/trained\_model.pth}, \texttt{results/ant\_training\_trace.json}, and \texttt{results/config\_resolved.json}, matching the expected pattern of checkpointed training and reproducible configuration logging.
When reference evidence is removed, the paper obligations remain visible, but the run has less support for file layout, data-handling conventions, experiment wiring, and artifact naming.
This case provides a concrete trace of why the reference-evidence channel can matter even when the target paper already names the high-level task.

\subsection{Additional Ablation Patterns}

The ablation pattern is not driven by a single paper.
The excerpts below are illustrative traces from the full Gemini repositories, not additional scoring criteria.
They show how the contract is materialized as file-owned implementation choices rather than remaining only in planning prose.

\begin{repocode}{Sample-specific Masks: method and ablation factory excerpt}
BASELINES = ["PAD", "NARROW", "MEDIUM", "FULL"]
ABLATIONS = ["ONLY_delta", "ONLY_f_mask",
             "SINGLE_CHANNEL_f_mask_s", "OURS"]

def get_method_config(method_name: str) -> MethodConfig:
    if method_name == "ONLY_delta":
        return MethodConfig(name="ONLY_delta", use_delta=True,
                            use_f_mask=False)
    elif method_name == "ONLY_f_mask":
        return MethodConfig(name="ONLY_f_mask", use_delta=False,
                            use_f_mask=True)
    elif method_name in ["ours", "OURS", "SMM"]:
        return MethodConfig(name="OURS", use_delta=True,
                            use_f_mask=True, mask_channels=3)
    elif method_name in ["PAD", "NARROW", "MEDIUM", "FULL"]:
        return MethodConfig(name=method_name, use_delta=True,
                            use_f_mask=False)
\end{repocode}

\begin{repocode}{Bridging Data Gaps: training artifact excerpt}
for step in range(num_iterations):
    loss_val = train_ant_step(batch, config, model, adaptor,
                              classifier, optimizer)
    trace.append({"step": step, "loss": loss_val})

os.makedirs("checkpoints", exist_ok=True)
os.makedirs("results", exist_ok=True)
torch.save(adaptor.state_dict(), "checkpoints/adaptor.pth")
torch.save(model.state_dict(), "checkpoints/trained_model.pth")

with open("results/ant_training_trace.json", "w") as f:
    json.dump(trace, f, indent=2)
with open("results/config_resolved.json", "w") as f:
    json.dump(config, f, indent=2)
\end{repocode}

\begin{table*}[t]
\centering
\small
\setlength{\tabcolsep}{3pt}
\begin{tabular}{lrrrp{0.46\textwidth}}
\toprule
Paper & Full & w/o impl. & w/o ref. & Case role \\
\midrule
Sample-specific Masks & 53.8 & 21.3 & 28.0 & Clear two-channel benefit: the full run preserves SMM-specific optimization, mask generation, interpolation, baselines, and ablations. \\
BAM & 48.5 & 30.8 & 20.7 & Large positive case: both removed-channel settings score substantially below the full contract, with the largest gap for reference evidence. \\
Bridging Data Gaps & 50.6 & 38.4 & 13.9 & Reference-heavy case: paper obligations remain useful, but removing reference evidence sharply reduces concrete protocol and repository-structure recovery. \\
PINN & 61.3 & 53.3 & 52.2 & Standardized-prior case: all variants remain strong because PINN objectives and PDE-loss structure are common, but the full contract is still best. \\
What Will My Model Forget & 16.8 & 10.7 & 8.4 & Scale and coverage stress case: the task has many long-tail details, and the full contract still leads both ablations. \\
LCA-on-the-Line & 41.8 & 16.1 & 33.8 & Boundary case: no-reference keeps a visible checklist for ObjectNet/LCA details, while the full run remains the strongest setting. \\
\bottomrule
\end{tabular}
\caption{Representative Gemini ablation cases. The case role column summarizes inspection observations alongside the score columns. ``w/o impl.'' removes the implementation-requirement channel; ``w/o ref.'' removes the reference-evidence channel.}
\label{tab:additional_ablation_cases}
\end{table*}

\begin{figure*}[t]
  \centering
  \includegraphics[width=0.88\textwidth]{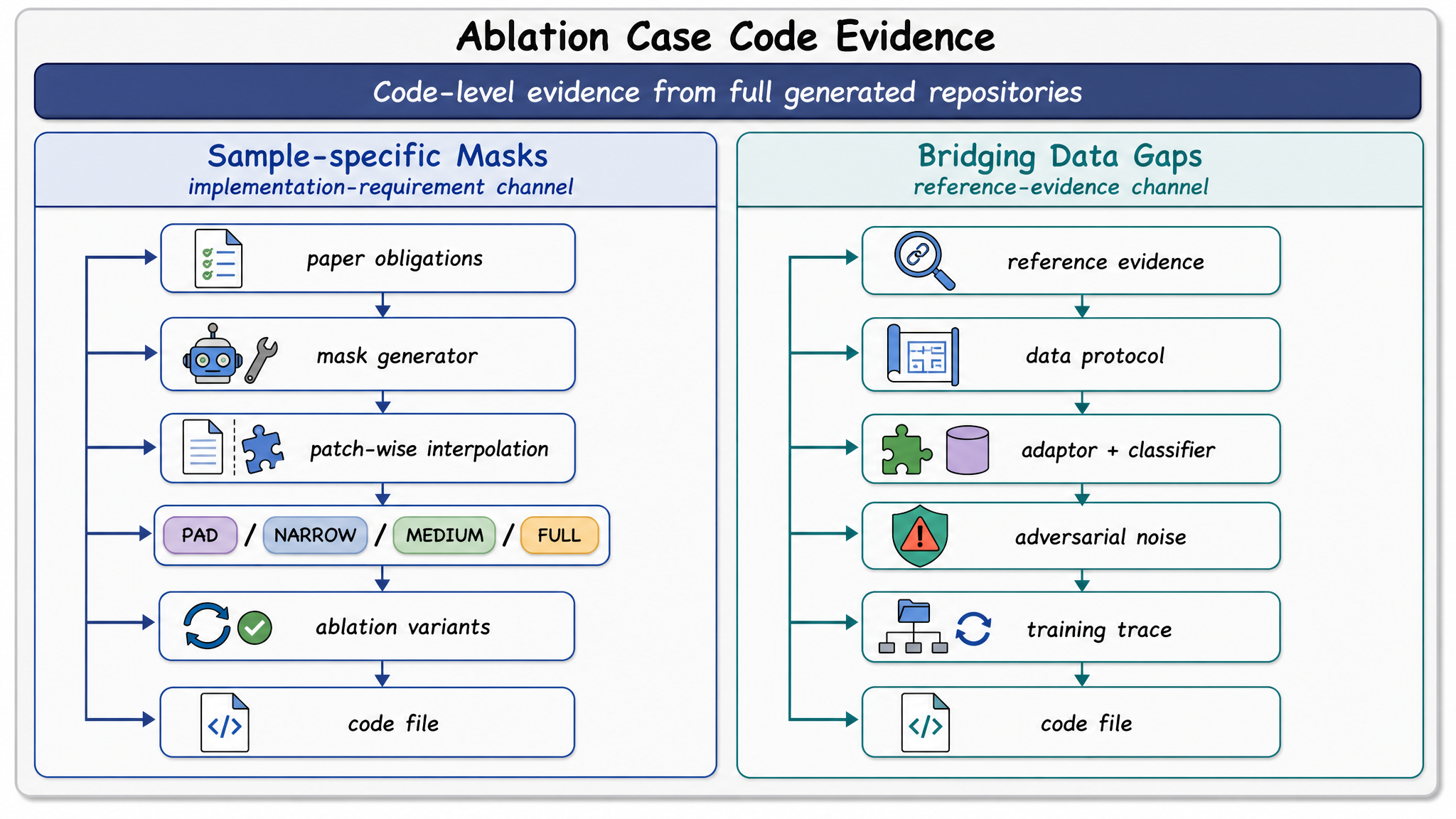}
  \caption{Code-level evidence for two representative ablation cases. The full generated repositories expose different concrete surfaces: Sample-specific Masks relies on implementation-requirement preservation for method-specific masks, interpolation, baselines, and ablations, while Bridging Data Gaps relies more heavily on reference evidence for protocol and training-artifact wiring.}
  \label{fig:ablation_case_code_evidence}
\end{figure*}

\paragraph{Large two-channel gains.}
Sample-specific Masks and BAM are representative qualitative traces covering large two-channel gains.
In Sample-specific Masks, the full run keeps the paper's required surfaces connected: the mask generator, interpolation from a low-resolution mask to the input image size, joint optimization of the shared pattern and mask-generator parameters, PAD/NARROW/MEDIUM/FULL baselines, and ablations over shared pattern and mask channels.
Removing implementation requirements disrupts this preservation path, so the repository retains runnable scaffolding but covers fewer paper-specific obligations.
Removing reference evidence keeps the obligations but reduces the file and experiment organization needed to expose baselines and ablation routes in a concrete implementation.
BAM shows a similar pattern at a coarser level: the full score is 48.5, while no-implementation and no-reference fall to 30.8 and 20.7, respectively.
This indicates that the reference-evidence channel recovers practical implementation and evaluation structure, while the implementation-requirement channel keeps paper-specific modeling and protocol obligations explicit.

\paragraph{Reference-sensitive cases.}
Bridging Data Gaps is useful because the two ablations degrade asymmetrically.
The full run reaches 50.6, the no-implementation run remains at 38.4, and the no-reference run drops to 13.9.
The artifact trace shows that the implementation-requirement channel preserves high-level tasks and expected outputs, while reference evidence supplies much of the concrete protocol surface: file layout, scripts, data handling conventions, and experiment wiring.
This is the kind of paper where related repositories supply implementation texture that is not explicit enough in the paper text alone.

\paragraph{Standardized tasks.}
PINN is a high-scoring but smaller-margin case.
The full run obtains 61.3, while the two ablations remain at 53.3 and 52.2.
This pattern is consistent with the channel hypothesis.
PINN-style repositories have strong public priors: collocation points, PDE residual losses, boundary losses, optimizer loops, and diagnostic plots are common implementation patterns.
As a result, Gemini can recover a reasonable repository even with one channel removed.
The full contract still gives the best score because it binds these common priors to the paper-specific objective, PDE setup, loss-landscape diagnostics, and expected artifacts.

\paragraph{Scale and coverage stress.}
What Will My Model Forget is a scale-sensitive case for the Gemini backbone.
The full setting scores 16.8 and both ablations are lower.
The target has broad long-tail coverage requirements and many benchmark-specific details, so this case separates channel utility from backbone capacity while preserving the same ordering pattern.

\subsection{Why Ablation Gaps Vary}
\label{sec:ablation_gap_heterogeneity}

The per-paper ablation gaps are heterogeneous, and this variation is itself informative.
For this discussion, we define the reference-evidence gain as the difference between the full Gemini score and the score without reference evidence, and the implementation-requirement gain as the difference between the full Gemini score and the score without implementation requirements.
The largest reference-evidence gains occur on Bridging Data Gaps (+36.7), Adaptive Pruning (+35.7), Test-Time Adaptation (+31.1), FTRL (+28.4), All-in-One (+27.9), and BAM (+27.8).
The largest implementation-requirement gains occur on Mechanistic Understanding (+43.5), Sample-Specific Masks (+32.5), Adaptive Pruning (+27.6), FTRL (+26.7), LCA-on-the-Line (+25.7), and All-in-One (+23.6).
These rankings suggest that the two channels help for related but distinguishable reasons.

\paragraph{Reference transferability.}
Reference evidence helps most when external artifacts provide transferable implementation structure: dataset protocols, baseline organization, metric wiring, experiment scripts, configuration conventions, or repository layout.
This is broader than simply counting cloned repositories.
For example, BBOX has multiple reference or dataset repositories in the preparation artifacts, yet its reference-evidence gain is small (+1.0), showing that repository count alone is not the key factor.
Conversely, Bridging Data Gaps shows a large reference-evidence gain even with a compact prepared reference-repository list, because the useful signal is whether reference evidence can be fused into concrete protocol and repository-structure decisions.
In reference-sensitive cases, removing this channel leaves the paper obligations visible but makes the generated repository less able to recover executable data handling, experiment wiring, and comparison surfaces.

\paragraph{Requirement hardness.}
The implementation-requirement channel helps most when a paper's contribution is compact but implementation-critical: a small set of named mechanisms, constraints, losses, ablations, metrics, or artifacts determines whether the repository is faithful.
Sample-Specific Masks is a representative case: the decisive obligations include mask generation, interpolation from a low-resolution mask to image size, joint optimization of shared patterns and mask-generator parameters, PAD/NARROW/MEDIUM/FULL baselines, and ablations over shared-pattern and mask channels.
Mechanistic Understanding, Adaptive Pruning, FTRL, and LCA-on-the-Line exhibit the same pattern at different scales: without explicit implementation requirements, the model can still produce plausible scaffolding, but it tends to omit or dilute the paper-specific mechanisms and protocol obligations that decide the score.

\paragraph{Small-gap cases.}
Small ablation gaps do not necessarily imply that the removed channels are irrelevant.
They usually fall into one of three categories.
First, some papers have strong public implementation priors; PINN is the clearest example, where collocation points, PDE residual losses, boundary losses, optimizer loops, and diagnostic plots are common enough that both ablations remain strong.
Second, some papers have compressed gaps because all variants face the same central implementation bottleneck; BBOX is a case where the three Gemini variants remain close.
Third, some tasks are scale or coverage stress cases, such as What Will My Model Forget, where the full contract still leads but the target contains many long-tail requirements.

Overall, the ablation gaps are best interpreted as indicators of where each channel provides leverage.
Reference evidence is most valuable when external artifacts are transferable into implementation structure, while implementation requirements are most valuable when fidelity depends on non-obvious paper-specific obligations.
Small gaps are therefore ambiguous: they may reflect strong public priors, limited reference transferability, or shared bottlenecks across variants, rather than evidence that the channels are unnecessary.

\section{Prompting and Structured Outputs}
\label{sec:prompts}

The production prompts use a shared system-user pattern.
The current pipeline uses schema-constrained JSON for planning stages, deterministic evidence grounding for package-level evidence bundles, and repository-edit prompts for generation and repair.
We therefore report compact schema excerpts rather than the full prompt text.
The released repository contains the full prompt templates and structured-output schemas.
\begin{promptbox}{Prepare: ExtractedUnit schema excerpt}
{
  "unit_id": "unit_002",
  "type": "method | protocol | claim | artifact | task",
  "statement": "paper-specific implementation obligation",
  "hypothesis": "what claim or mechanism this unit tests",
  "decision_value": "what implementation decision depends on it",
  "paper_evidence": ["source paper/addendum excerpt"],
  "source_paragraph_ids": ["chunk_006_01"],
  "reference_query_terms": ["optional explicit method or repository cue"],
  "implementation_surfaces": ["model_or_method", "metric_formula"],
  "code_obligations": ["concrete executable obligation"],
  "runtime_interfaces": ["CLI/function/config surface"],
  "expected_artifacts": ["results/metrics.json"],
  "suggested_module_kinds": ["training_loop", "artifact_writer"],
  "implementation_notes": ["inventory or scope notes"],
  "tags": ["method", "evaluation"]
}
\end{promptbox}
\begin{promptbox}{Plan: contract and file-planning schema excerpt}
{
  "work_package": {
    "work_package_id": "wp_foa_core",
    "owned_unit_ids": ["unit_002", "unit_003"],
    "inventories": {
      "obligation_matrix": ["Method -> file path"],
      "method_inventory": ["named method or baseline"],
      "parameter_inventory": ["paper-visible values"]
    },
    "scope_boundary": {
      "preserve": ["unit-level obligations that must survive"],
      "implementation_focus": ["active routes and artifacts"]
    },
    "method_obligations": ["what code must expose"]
  },
  "file_plan": {
    "target_file": "methods/foa.py",
    "work_package_id": "wp_foa_core",
    "owned_units": ["unit_002", "unit_003"],
    "reference_ids": ["ref or paper chunk ids"],
    "implementation_surfaces": ["model_or_method"],
    "defines_symbols": ["FOA", "CMAES"],
    "calls_symbols": ["fitness_function"],
    "writes_artifacts": ["results/adaptation_trace.json"],
    "scope_boundary": {"preserve": ["file-owned obligations"]},
    "generation_prompt": "file-local implementation instruction",
    "review_points": ["semantic and route-closure checks"]
  }
}
\end{promptbox}
\newpage
\begin{promptbox}{Evidence grounding and repair schema excerpt}
{
  "evidence_bundle": {
    "work_package_id": "wp_foa_core",
    "focus": "implementation duty",
    "owned_unit_ids": ["unit_002", "unit_003"],
    "evidence_links": [
      {"unit_id": "unit_002", "ref_id": "ref_001", "file_path": "repo/file.py",
       "snippet_preview": "...", "why_relevant": "...", "confidence": 0.0}
    ],
    "context_summary": ["paper or reference evidence summary"],
    "grounding_status": "grounded | self_contained | weak | ungrounded"
  },
  "repair_ticket": {
    "failure_type": "semantic | runtime | integration",
    "required_fix_targets": ["methods/foa.py"],
    "allowed_changes": ["targeted repair scope"],
    "forbidden_changes": ["requirements already preserved"],
    "next_fix_scope": ["files to patch"]
  }
}
\end{promptbox}

\section{Additional Algorithmic Details}
\label{sec:algorithmic_details}

\subsection{Algorithmic Summary}
\label{sec:algorithmic_summary}

\begin{algorithm}[H]
\caption{\system{} pipeline}
\label{alg:pipeline}
\begin{algorithmic}[1]
\REQUIRE Paper $\mathcal{P}$, request $r$, references $\mathcal{R}$, rounds $T$
\ENSURE Repository $\hat{\mathcal{C}}$
\STATE $\mathcal{S}\leftarrow\mathrm{ChunkPaper}(\mathcal{P})$
\STATE $\mathcal{Q}\leftarrow\mathrm{SurveyRepos}(\mathcal{R})$
\STATE $\mathcal{U}\leftarrow\mathrm{ExtractRequirements}(\mathcal{S})$
\STATE $\mathcal{U}\leftarrow\mathrm{Deduplicate}(\mathcal{U})$
\STATE $\mathcal{W}\leftarrow\mathrm{PlanWorkPackages}(\mathcal{U})$
\STATE $\mathrm{CheckPreserved}(\mathcal{U},\mathcal{W})$
\STATE $\mathcal{E}\leftarrow\mathrm{CollectContentEvidence}(\mathcal{W},\mathcal{Q})$
\STATE $\mathcal{Z}\leftarrow\mathrm{CollectStructureEvidence}(\mathcal{Q})$
\STATE $\mathcal{G}\leftarrow\mathrm{BuildContract}(\mathcal{U},\mathcal{W},\mathcal{E},\mathcal{Z})$
\STATE $\mathcal{F}\leftarrow\mathrm{PlanFiles}(\mathcal{G})$
\FOR{$f\in\mathrm{TopologicalOrder}(\mathcal{F})$}
  \STATE $\hat{f}\leftarrow\mathrm{GenerateFile}(f,\mathcal{G},\mathcal{E})$
\ENDFOR
\FOR{$t=1$ to $T$}
  \STATE $v_q\leftarrow\mathrm{RequirementReview}(\hat{\mathcal{C}},\mathcal{U})$
  \STATE $v_r\leftarrow\mathrm{RuntimeValidation}(\hat{\mathcal{C}})$
  \IF{$v_q.\mathrm{passed}\land v_r.\mathrm{passed}$} \STATE \textbf{break} \ENDIF
  \STATE $\tau\leftarrow\mathrm{RepairTicket}(v_q,v_r,\mathcal{E})$
  \STATE $\hat{\mathcal{C}}\leftarrow\mathrm{ApplyRepair}(\hat{\mathcal{C}},\tau)$
\ENDFOR
\RETURN $\hat{\mathcal{C}}$
\end{algorithmic}
\end{algorithm}

\subsection{Repair Loop}
\label{sec:repair_loop_appendix}

\begin{algorithm}[H]
\caption{Contract-preserving repair loop}
\label{alg:repair_loop}
\begin{algorithmic}[1]
\REQUIRE Repository $\hat{\mathcal{C}}$, requirements $\mathcal{U}$, contracts $\mathcal{G}$, evidence $\mathcal{E}$, budget $T$
\ENSURE Patched repository $\hat{\mathcal{C}}$
\FOR{$t=1$ to $T$}
  \STATE $s\leftarrow\mathrm{ReviewRequirements}(\hat{\mathcal{C}},\mathcal{U},\mathcal{G})$
  \STATE $r\leftarrow\mathrm{RunValidation}(\hat{\mathcal{C}})$
  \IF{$s.\mathrm{passed}\land r.\mathrm{passed}$}
    \STATE \textbf{break}
  \ENDIF
  \STATE $\tau\leftarrow\mathrm{BuildRepairTicket}(s,r,\mathcal{G},\mathcal{E})$
\STATE $\tau_{\mathrm{ban}}\leftarrow\mathrm{PreservedReqs}(s)$
  \STATE $\hat{\mathcal{C}}\leftarrow\mathrm{PatchAffectedFiles}(\hat{\mathcal{C}},\tau)$
\ENDFOR
\RETURN $\hat{\mathcal{C}}$
\end{algorithmic}
\end{algorithm}

\section{Per-Paper Scores and Generated Repository Sizes}
\label{sec:appendix_full_scores}

This section gives the complete per-paper scores and generated repository sizes behind the full and ablated runs reported in the main results.

\begin{table*}[t]
\centering
\small
\setlength{\tabcolsep}{3pt}
\begin{tabular*}{\textwidth}{@{\extracolsep{\fill}}>{\raggedright\arraybackslash}p{0.45\textwidth}rrrrrrrr@{}}
\toprule
\multirow{2}{*}{Paper} & \multicolumn{2}{c}{Claude} & \multicolumn{2}{c}{Gemini} & \multicolumn{2}{c}{w/o ref.} & \multicolumn{2}{c}{w/o req.} \\
\cmidrule(lr){2-3}\cmidrule(lr){4-5}\cmidrule(lr){6-7}\cmidrule(l){8-9}
& Score & Files & Score & Files & Score & Files & Score & Files \\
\midrule
Adaptive Pruning & 67.8 & 46 & 41.9 & 92 & 6.2 & 23 & 14.3 & 24 \\
All in One & 76.9 & 77 & 51.6 & 48 & 23.7 & 16 & 28.0 & 36 \\
BAM & 61.8 & 37 & 48.5 & 47 & 20.7 & 18 & 30.8 & 38 \\
BBOX & 86.4 & 280 & 15.6 & 70 & 14.6 & 15 & 14.2 & 75 \\
Bridging Data Gaps & 66.0 & 151 & 50.6 & 23 & 13.9 & 17 & 38.4 & 19 \\
FRE & 76.8 & 23 & 26.9 & 32 & 12.7 & 14 & 25.1 & 20 \\
FTRL & 63.9 & 28 & 40.1 & 71 & 11.7 & 23 & 13.4 & 51 \\
LBCS & 65.8 & 40 & 42.5 & 51 & 34.6 & 24 & 23.2 & 22 \\
LCA on the Line & 64.9 & 30 & 41.8 & 22 & 33.8 & 18 & 16.1 & 15 \\
Mech. Understanding & 75.7 & 201 & 48.1 & 26 & 25.9 & 14 & 4.6 & 20 \\
PINN & 81.5 & 163 & 61.3 & 35 & 52.2 & 14 & 53.3 & 18 \\
RICE & 75.1 & 34 & 31.0 & 81 & 19.9 & 16 & 27.8 & 25 \\
Robust CLIP & 63.0 & 51 & 29.5 & 19 & 15.7 & 14 & 25.6 & 20 \\
Sample-specific Masks & 82.0 & 80 & 53.8 & 50 & 28.0 & 17 & 21.3 & 33 \\
SAPG & 86.5 & 141 & 29.0 & 25 & 21.5 & 16 & 24.2 & 24 \\
Sequential Neural Score Estimation & 78.0 & 30 & 45.7 & 14 & 21.5 & 13 & 42.6 & 26 \\
Stay on Topic with Classifier-Free Guidance & 58.5 & 31 & 25.3 & 21 & 21.2 & 20 & 22.2 & 27 \\
Stochastic Interpolants & 80.8 & 73 & 39.1 & 25 & 21.5 & 19 & 29.9 & 24 \\
Test-Time Model Adaptation & 75.6 & 49 & 54.9 & 20 & 23.8 & 12 & 47.1 & 26 \\
What Will My Model Forget & 87.6 & 23 & 16.8 & 22 & 8.4 & 12 & 10.7 & 19 \\
\midrule
\textbf{Mean} & \textbf{73.7} & \textbf{79.4} & \textbf{39.7} & \textbf{39.7} & \textbf{21.6} & \textbf{16.8} & \textbf{25.6} & \textbf{28.1} \\
\bottomrule
\end{tabular*}
\caption{Per-paper scores and generated repository sizes for our full and ablated runs. File counts come from the corresponding run metadata and are reported as process metadata rather than scoring inputs.}
\label{tab:appendix_our_scores_files}
\end{table*}

\section{Token, Time, and Cost Diagnostics}
\label{sec:usage_diagnostics}

Tables~\ref{tab:appendix_usage_summary} and~\ref{tab:appendix_usage_by_paper} report usage metadata for interpreting the reproduction process.
Scoring-token counts come from PaperBench judge summaries; when a judge summary stores token usage at leaf level, we sum the recoverable per-leaf metadata from the same graded task tree.
Pipeline-token counts, wall-clock time, and agent-call counts use exported run logger summaries when available.
Runtime metadata is exported at different granularity across settings; the Claude pipeline-token cell is estimated from matched run logs and is reported only as process metadata, while hours and calls use the recoverable run metadata directly.
For the Gemini family, the full run uses more pipeline tokens than either ablation, which is expected because the full setting carries both implementation requirements and reference evidence through planning and repair.
Wall-clock time varies with target-paper difficulty and available runtime metadata.

\begin{table*}[t]
\centering
\small
\setlength{\tabcolsep}{3pt}
\begin{tabular}{@{}lrrrrrr@{}}
\toprule
Setting & Papers & Score & Judge tok. & Pipe tok. & Hours & Calls \\
\midrule
ReproAgent / Claude & 20 & 73.7 & 12,868,729 & 17,833,795 & 9.3 & 35.4 \\
ReproAgent / Gemini full & 20 & 39.7 & 6,243,588 & 17,459,007 & 5.6 & 11.7 \\
Gemini w/o reference evidence & 20 & 21.6 & 6,582,787 & 4,034,007 & 1.5 & 13.8 \\
Gemini w/o implementation requirements & 20 & 25.6 & 4,668,761 & 4,429,321 & 1.3 & 12.6 \\
\bottomrule
\end{tabular}
\caption{Usage metadata for our runs. Judge-token cells use root-level or summed leaf metadata depending on how the PaperBench summary stores usage; pipeline-token cells use exported run metadata or the estimate described above.}
\label{tab:appendix_usage_summary}
\end{table*}

\begin{table*}[t]
\centering
\small
\setlength{\tabcolsep}{1.2pt}
\begin{tabular}{@{}>{\raggedright\arraybackslash}p{0.24\textwidth}rrrrrrr@{}}
\toprule
Paper & Claude judge & Gemini judge & Gemini pipe & No-ref judge & No-ref pipe & No-req judge & No-req pipe \\
\midrule
Adaptive pruning & 5,423,412 & 2,868,933 & 23,047,479 & 1,980,736 & 2,236,269 & 3,341,264 & 5,919,631 \\
All-in-one & 5,538,759 & 3,780,967 & 9,264,024 & 3,429,304 & 3,936,893 & 2,250,662 & 2,630,553 \\
BAM & 30,766,863 & 7,768,083 & 16,097,296 & 8,827,504 & 4,969,452 & 1,892,690 & 15,027,467 \\
BBOX & 8,567,577 & 774,669 & 58,268,936 & 4,903,571 & 4,131,573 & 4,373,037 & 4,509,759 \\
Bridging data gaps & 4,746,750 & 2,300,086 & 8,820,069 & 1,864,915 & 4,264,829 & 2,412,905 & 4,406,997 \\
FRE & 37,387,691 & 13,780,748 & 8,425,041 & 10,006,741 & 2,975,272 & 12,737,673 & 3,925,572 \\
FTRL & 11,530,535 & 12,388,835 & 23,625,213 & 3,976,499 & 5,255,585 & 291,099 & 4,136,571 \\
LBCS & 35,890,914 & 17,312,480 & 9,390,925 & 20,577,036 & 5,455,156 & 9,045,550 & 6,665,934 \\
LCA on the line & 25,328,353 & 613,700 & 8,697,850 & 18,609,830 & 4,585,214 & 4,880,558 & 3,644,005 \\
Mech. understanding & 2,176,841 & 1,589,700 & 7,923,157 & 1,340,680 & 3,134,806 & 1,383,850 & 4,118,063 \\
PINN & 7,360,909 & 5,691,891 & 13,620,939 & 7,053,399 & 4,120,951 & 6,536,328 & 4,348,687 \\
RICE & 16,389,279 & 1,550,040 & 74,534,103 & 6,929,485 & 3,739,967 & 5,148,614 & 3,261,044 \\
Robust CLIP & 5,588,405 & 2,147,322 & 16,285,677 & 3,131,589 & 3,445,426 & 2,689,292 & 2,283,182 \\
Sample-specific masks & 9,405,505 & 2,230,104 & 18,789,684 & 2,932,425 & 4,813,964 & 604,261 & 2,426,577 \\
SAPG & 4,553,396 & 3,507,987 & 8,202,509 & 3,339,036 & 3,793,693 & 2,585,659 & 4,160,601 \\
Sequential NSE & 3,293,860 & 2,661,812 & 6,486,552 & 2,075,601 & 4,250,141 & 2,723,479 & 5,711,514 \\
Stay on topic & 3,206,265 & 1,684,783 & 13,880,313 & 2,100,836 & 4,674,944 & 159,171 & 2,992,827 \\
Stochastic interpolants & 3,641,790 & 2,171,919 & 8,435,217 & 1,364,707 & 4,452,320 & 1,731,008 & 1,227,333 \\
Test-time adaptation & 7,102,606 & 3,326,170 & 7,495,412 & 3,511,711 & 3,539,717 & 336,977 & 5,935,940 \\
What will my model forget & 29,474,870 & 36,721,521 & 7,889,747 & 23,700,128 & 2,903,969 & 28,251,151 & 1,254,158 \\
\bottomrule
\end{tabular}
\caption{Per-paper token metadata. ``Judge'' is PaperBench grading-token usage, with leaf metadata summed for summaries that store usage at leaf level; ``pipe'' is generation-token usage from exported run metadata when available.}
\label{tab:appendix_usage_by_paper}
\end{table*}

\section{Full Pipeline Stage Sequence}
\label{sec:full_pipeline}

\system{} comprises 15 sub-stages organized into four phases.
Table~\ref{tab:pipeline_stages} lists every stage with its phase assignment and a brief description.

\begin{table*}[t]
\centering
\small
\begin{tabular}{@{}p{0.12\textwidth}p{0.23\textwidth}p{0.58\textwidth}@{}}
\toprule
\textbf{Phase} & \textbf{Stage} & \textbf{Engineering role} \\
\midrule
Prepare  & Input normalization   & Normalize the target, task type, entities, and explicit constraints. \\
Prepare  & Paper chunking        & Segment paper text into section-level chunks with overflow handling. \\
Prepare  & Requirement extraction & Extract implementation requirements that need concrete code, artifacts, interfaces, or validation targets. \\
Prepare  & Reference acquisition & Collect reference repositories and survey file trees, README files, and likely reusable modules. \\
\midrule
Plan & Work-package planning & Group prepared requirements into implementation-oriented packages such as data, method, training, evaluation, and artifact writing. \\
Plan & Package reference evidence & Attach package-local content and structure evidence; this supports package decisions but does not decide the final file tree. \\
Plan & Reference selection & Select actionable reference repositories and evidence items for the current package needs. \\
Plan & Pipeline plan & Specify package dependencies, runnable routes, and ordering constraints. \\
Plan & Global contract synthesis & Freeze requirement ownership, evidence bindings, result targets, artifact paths, interfaces, validation checks, and cross-package dependencies. \\
Plan & Architecture planning & Decide file tree, stack, entrypoints, config surfaces, dependency rules, and package-to-file layout. \\
Plan & Package-file planning & Refine architecture into per-file contracts with owned requirements, evidence, interfaces, artifacts, and validation hooks. \\
\midrule
Generate & Canonical representation synthesis & Consolidate requirements, packages, contracts, architecture, file plans, evidence links, and validator expectations. \\
Generate & Local file generation & Generate source code file by file following file contracts and dependency ordering. \\
\midrule
Repair & Repair validation & Validate through requirement review, contract checks, preflight checks, and sandboxed runtime execution. \\
Repair & Repair regeneration & Build a targeted repair plan and patch affected files while preserving requirements that already pass. \\
\bottomrule
\end{tabular}
\caption{Complete stage sequence of the \system{} pipeline. Stages are executed in order; within-phase stages may share intermediate artifacts.}
\label{tab:pipeline_stages}
\end{table*}

The Repair phase operates as an iterative loop with a fixed budget within each backbone setting.
In each iteration, the system executes validation, diagnosis, and regeneration in sequence.
The loop terminates when all validation checks pass or the budget is exhausted.

\subsection{Process Trace and Readiness}
\label{sec:process_trace_readiness}

Each run records intermediate artifacts for auditing each phase, not only the final repository and score.
The stable audit records include prepared units, reference-repository records, work-package plans, contract assertions, file-planning artifacts, validation reports, stage status, quality status, and usage summaries.
Table~\ref{tab:process_trace_diagnostics} summarizes coarse audit statistics from the 20 Gemini full runs.

\begin{table*}[t]
\centering
\small
\begin{tabular}{@{}llrrr@{}}
\toprule
Stage & Audit statistic & Mean & Median & Range \\
\midrule
Prepare & Extracted implementation units per paper & 33.8 & 34.0 & 28--38 \\
Prepare & Candidate reference repositories requested per paper & 24.6 & 25.0 & 10--30 \\
Prepare & Prepared reference repositories per paper & 1.1 & 0.0 & 0--9 \\
Prepare & Implementation units grounded by references per paper & 6.2 & 0.0 & 0--29 \\
Plan & Work packages per paper & 19.6 & 21.5 & 3--38 \\
Plan & Contract assertions per paper & 6.9 & 7.0 & 5--10 \\
Generate & Generated files per paper & 35.5 & 26.5 & 14--80 \\
Runtime metadata & Pipeline agent calls per paper & 11.7 & 6.0 & 3--29 \\
\bottomrule
\end{tabular}
\caption{Coarse audit statistics from the 20 Gemini full runs. These logged counts summarize the scale of preparation, planning, generation, and runtime interaction recorded during execution; they are not inputs to PaperBench judging.}
\label{tab:process_trace_diagnostics}
\end{table*}

These logged counts make the generation process more concrete.
Prepare extracts implementation units from the paper itself rather than from the PaperBench rubric, then attempts to ground those units with related repositories when actionable evidence is available.
Prepared reference repositories are concentrated in a subset of papers, reflecting the benchmark's mix of papers with different amounts of reusable prior-work structure.
Plan then turns prepared units into work packages and contract assertions, which are contract-shaping objects rather than file-count targets.
Generate materializes the file plans into repositories, and Repair audits requirement, trace, artifact, integration, and runtime checks.

These logged counts are not inputs to PaperBench judging.
All reported Gemini full runs completed repository generation and were submitted to the PaperBench judge.
We report them alongside PaperBench scores to make the pipeline trace auditable.

\end{document}